\documentclass{article}
\usepackage{spconf,amsmath,graphicx,hyperref}

\usepackage{booktabs}   
\usepackage{makecell}   

\usepackage{booktabs}
\usepackage{pifont}
\usepackage{multirow}
\usepackage{enumitem}
\usepackage{caption}
\usepackage{eso-pic}

\usepackage[table]{xcolor}

\newcommand{\cmark}{\ding{51}}
\newcommand{\xmark}{\ding{55}}

\title{OmniRoute: Mapping Temporal Semantic Evidence to Audio-Visual Token Budgets for Efficient Omnimodal Large Language Models}
\name{
Yuchen Deng$^{1,2,*}$,
Zidang Cai$^{1,*}$,
Feidiao Yang$^{2}$,
Yufei Wang$^{1,2}$,
Jie Wang$^{1,2}$,
Hai-Tao Zheng$^{1,2}$,
Yuxing Han$^{1,\dagger}$%
\thanks{%
\fontsize{9}{10.8}\selectfont
$^{*}$Equal contribution.\quad
$^{\dagger}$Corresponding author.
}
}

\address{
$^{1}$Shenzhen International Graduate School, Tsinghua University, China\\
$^{2}$Pengcheng Laboratory, China
}

\begin{document}
\ninept
\maketitle

\AddToShipoutPictureFG*{%
  \AtTextUpperLeft{%
    \put(0,\LenToUnit{10mm}){%
      \parbox[t]{\textwidth}{%
        \normalfont\fontsize{9}{10.8}\selectfont
        \raggedright
        This work has been submitted to the IEEE for possible publication.\\
        Copyright may be transferred without notice, after which this version may no longer be accessible.
      }%
    }%
  }%
}
\begin{abstract}
Omnimodal large language models (Omni-LLMs) encode audio and visual streams into temporally interleaved token sequences for multimodal reasoning. However, processing long audio-visual token sequences incurs substantial prefill costs. Existing compression methods have made progress, but often overlook temporal changes in audio-visual semantic relevance. Motivated by temporal variation and local continuity, we propose OmniRoute, a training-free, two-stage compression framework. First, Temporal Evidence-Guided Budgeting (TEGB) derives chunk-wise modality preferences and initial leading-modality budgets from semantic relevance and local content variation. Second, Budget-Constrained Semantic Compression (BCSC) compresses the leading modality and then calibrates the follower's retention target using the actual retained fraction. For video, it combines spatiotemporal grouping with query-guided selection; for audio, it selects tokens based on encoder attention and query relevance, then merges residual tokens into context anchors under visual guidance. Experiments on four representative benchmarks demonstrate a better trade-off between inference efficiency and performance than competitive baselines. The code and interface will be released to facilitate further research.

\end{abstract}
%
%

\section{Introduction}

Omnimodal large language models (Omni-LLMs)~\cite{xu2025qwen25omnitechnicalreport} extend multimodal LLMs~\cite{chen2024internvl,li2025videochat} toward unified understanding of text, vision, and audio, with applications in multimedia content analysis and embodied intelligence. Representative Omni-LLMs, such as Qwen2.5-Omni, encode audio and visual streams into a temporally aligned token sequence for reasoning over complementary evidence~\cite{ye2025omnivinci,tong2025interactiveomni}. 

However, long audio-visual token sequences incur substantial prefill costs due to quadratic attention complexity. Therefore, token compression has emerged as an important direction~\cite{xu2025qwen3,team2026qwen3,sun2026omnimem} for efficient Omni-LLMs inference, aiming to reduce computational overhead while maintaining model performance.

Existing methods compress multimodal tokens before prefill through cross-modal guidance and structure-aware selection, preserving cross-modal complementarity~\cite{tao2025omnizip,gong2025echoingpixels,deng2026omnirefine}. Other methods reduce model-internal redundancy by adapting token and KV-cache retention using layer-wise signals and cross-modal interactions~\cite{wangomnifit,jung2026fastav,jiang2026acckv}. Recent methods further incorporate query relevance into token selection to prioritize task-relevant evidence~\cite{yang2026omniselect,park2026omnidrop,chen2026avoc}. However, these methods often overlook temporal changes in audio-visual semantic relevance. Across 500 WorldSense videos analyzed using Qwen2.5-Omni-7B, audio-visual semantic relevance exhibits temporal variation and local continuity. This motivates us to account for temporal characteristics in compression design.

To this end, we propose \textbf{OmniRoute}, a training-free, two-stage framework for semantics-aware token compression in Omni-LLMs. First, \textbf{Temporal Evidence-Guided Budgeting (TEGB)} estimates chunk-wise modality preferences by comparing the query relevance of audio and video. It then sets initial retention budgets for the leading modality based on preference strength and visual content variation within and across chunks. Second, \textbf{Budget-Constrained Semantic Compression (BCSC)} performs chunk-wise compression guided by the leading modality. It compresses this modality according to its initial budget and then uses its actual retention rate to calibrate the follower's target retention rate. For video, it constructs a candidate mask through hierarchical spatial grouping and temporal grouping of similar regions, then applies query-guided selection to meet the retention target. For audio, it combines encoder-attention importance with query relevance to select tokens, then merges selected residual tokens into context anchors under visual guidance.

\begin{figure}[t]
  \centering
  \includegraphics[width=\columnwidth]
  {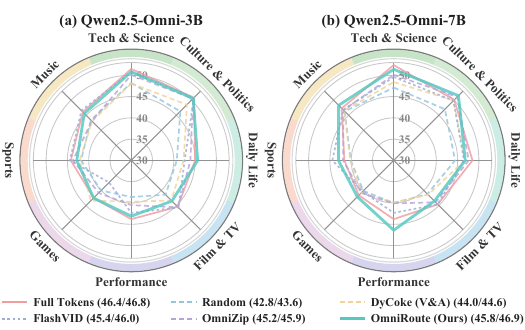}
  \vspace{-4mm}
  \caption{WorldSense accuracy using Qwen2.5-Omni-3B/7B.}
  \label{fig:worldsense_radar_comparison}
  \vspace{-4mm}
\end{figure}

Extensive experiments on WorldSense, AVUTBench, VideoMME, and DailyOmni show that OmniRoute consistently achieves a better efficiency-performance trade-off than strong baselines. As illustrated in Fig.~\ref{fig:worldsense_radar_comparison}, OmniRoute reaches 46.9\% accuracy on Qwen2.5-Omni-7B using 45\% of the tokens, exceeding the full-token baseline.

We summarize the contributions of this paper as follows.
\begin{itemize}[leftmargin=*, nosep]
    \item Motivated by temporal variation and local continuity in semantic relevance, we propose OmniRoute, a training-free, two-stage token compression framework for Omni-LLMs.
    \item We introduce TEGB for chunk-wise modality preferences and initial budgets reflecting semantic relevance and local content variation, and BCSC for modality-led semantic compression with cross-modal budget calibration using actual retention.
    \item Experiments demonstrate that OmniRoute better balances inference efficiency and model performance than baselines.
\end{itemize}

\section{Method}

\subsection{Preliminaries}
Existing methods may overlook temporal variation and local continuity in query relevance. For the $i$-th chunk, let $E_i^a$ and $E_i^v$ denote the mean cosine similarities between the mean query embedding $\mathbf{q}$ and the audio and visual token embeddings. Their contrast is
\begin{equation}
D_i = E_i^a-E_i^v.
\label{eq:contrast}
\end{equation}
Larger $D_i$ indicates higher query relevance of audio relative to vision, while the sequence $\{D_i\}_{i=1}^{n}$ tracks how this balance evolves across successive chunks.
Using Qwen2.5-Omni-7B, we analyze 500 randomly sampled WorldSense videos. For the $s$-th video with $n_s$ native chunks, we quantify dispersion using the scale-normalized interdecile range
\begin{equation}
V_s =
\frac{
Q_{0.9}\!\left(\{D_i^{(s)}\}_{i=1}^{n_s}\right)
-
Q_{0.1}\!\left(\{D_i^{(s)}\}_{i=1}^{n_s}\right)
}{
\operatorname*{median}_{1\leq i\leq n_s}\!\left(
|E_i^{a,(s)}|+|E_i^{v,(s)}|
\right)
+\varepsilon
}.
\label{eq:variation}
\end{equation}
We measure local continuity using
\begin{equation}
C_s =
\frac{
\frac{1}{n_s-1}
\sum_{i=1}^{n_s-1}
|D_{i+1}^{(s)}-D_i^{(s)}|
}{
\binom{n_s}{2}^{-1}
\sum_{1\leq i<j\leq n_s}
|D_i^{(s)}-D_j^{(s)}|
}.
\label{eq:continuity}
\end{equation}
Here, $Q_p$ denotes the $p$-quantile, and $\varepsilon=10^{-12}$ is a numerical stabilizer. A larger $V_s$ indicates greater within-video dispersion of $D_i^{(s)}$, while $C_s<1$ indicates that adjacent chunks differ less in $D_i^{(s)}$ than arbitrary chunk pairs.

\begin{figure}[t]
  \centering
  \includegraphics[width=\columnwidth]
  {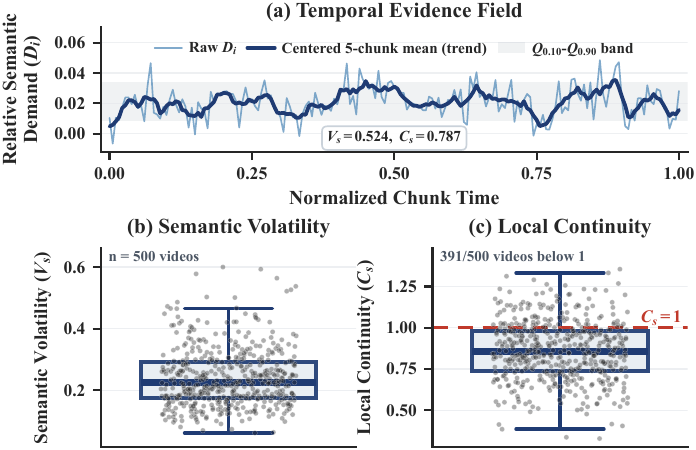}
  \vspace{-4mm}
  \caption{Temporal audio-visual evidence. (a) The $D_i$ trajectory and its centered five-chunk moving average for a WorldSense video. (b)-(c) Distributions of $V_s$ and $C_s$ over 500 videos.}
  \label{fig:motivating_analysis}
  \vspace{-4mm}
\end{figure}

As illustrated in Fig.~\ref{fig:motivating_analysis}(a), the $D_i$ trajectory of an example video exhibits recurring peaks and valleys while evolving coherently across neighboring chunks. Panels (b)-(c) summarize the results across 500 videos: $V_s$ ranges from $0.062$ to $0.600$ (median $0.225$), while $C_s$ is below $1$ for 391 videos ($78.2\%$) and averages $0.860$ (video-level bootstrap 95\% CI: $0.844$--$0.875$). Together, these results indicate that the query-conditioned audio-visual relevance contrast varies across chunks yet often remains locally continuous.

\subsection{Overview of OmniRoute}
\begin{figure*}[t]
  \centering
  \includegraphics[width=\textwidth]
  {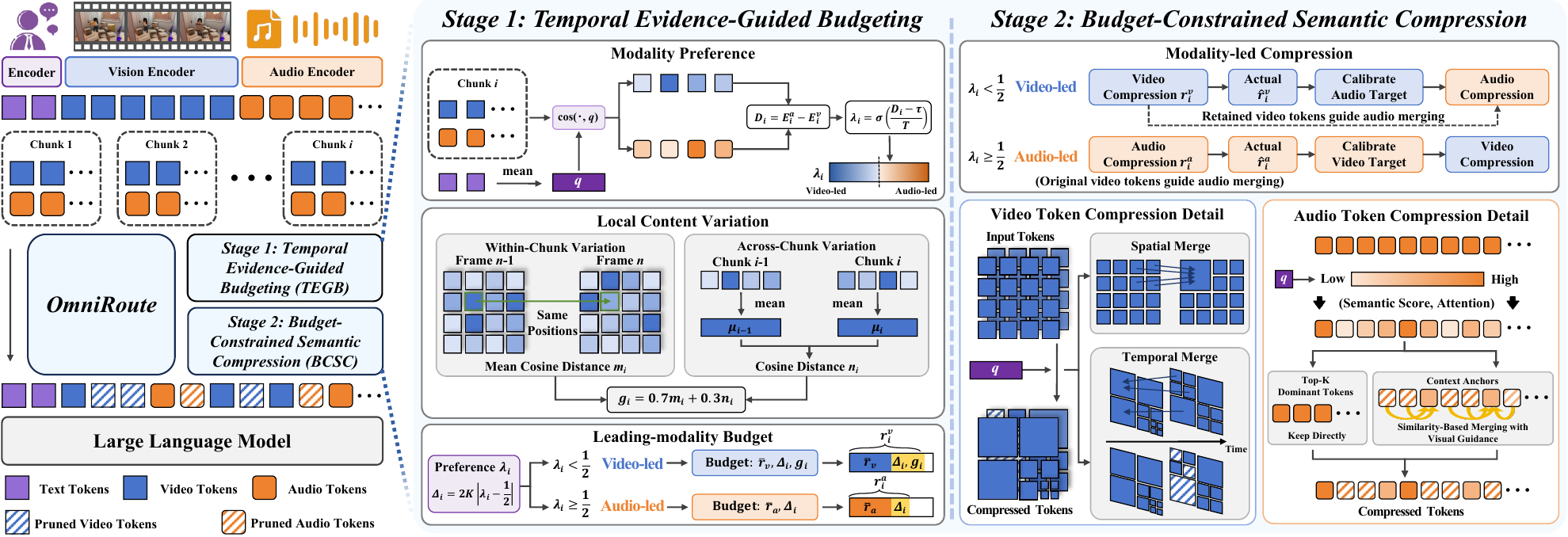}
  \vspace{-4.5mm}
  \caption{OmniRoute overview. TEGB derives chunk-wise modality preferences and leading-modality budgets from semantic relevance and local variation. BCSC calibrates follower targets using actual leading-modality retention. Video compression combines spatial and temporal grouping with query-guided selection; audio compression combines attention- and query-based selection with visually guided merging.}
  \label{fig:omniroute-overview}
  \vspace{-3.5mm}
\end{figure*}

OmniRoute is a training-free framework for semantics-aware audio-visual token compression in Omni-LLMs. As illustrated in Fig.~\ref{fig:omniroute-overview}, it operates before LLM prefill through two stages: Temporal Evidence-Guided Budgeting (TEGB) and Budget-Constrained Semantic Compression (BCSC).

\subsection{Temporal Evidence-Guided Budgeting}
TEGB estimates the modality preference for each chunk from audio-visual semantic evidence. It uses content changes within and between chunks, together with preference strength, to construct the leading modality’s retention budget.  

\noindent\textbf{Modality preference.}
Using the query-conditioned evidence contrast $D_i$ defined in Eq.~\eqref{eq:contrast}, we compute
\begin{equation}
\lambda_i=\sigma\!\left(\frac{D_i-\tau}{T}\right),
\label{eq:modality-preference}
\end{equation}
where $\sigma$ is the sigmoid, $\tau$ is the direction threshold, and the temperature $T>0$ controls preference sharpness. For each chunk, the leading modality is $\ell_i=a$ if $\lambda_i\geq 1/2$, and $\ell_i=v$ otherwise.

\noindent\textbf{Local content variation.}
We characterize visual changes at two temporal scales. Within chunk $i$, $m_i$ measures the mean cosine distance between visual features at corresponding spatial positions across consecutive time steps. 
Between adjacent chunks, we compare mean visual embeddings: $n_i=1-\cos(\boldsymbol\mu_i,\boldsymbol\mu_{i-1})$ for $i>1$, with $n_1=0$, where $\boldsymbol\mu_i$ averages visual token embeddings in chunk $i$. They adjust the initial visual retention rate:
\begin{align}
g_i &= 0.7m_i+0.3n_i,
\label{eq:local-content-variation}\\
r_{i,0}^{v}
&= \Pi_v\!\left(\overline r_v+\alpha(g_i-\overline g)\right),
\label{eq:local-content-budget}
\end{align}
where $\overline g=n^{-1}\sum_{j=1}^{n}g_j$ is the mean variation score over the $n$ chunks of the video, $\overline r_v$ is the base visual retention rate, $\alpha\geq0$ controls sensitivity, and $\Pi_v$ clips visual retention to its allowed range.

\noindent\textbf{Leading-modality budget.}
We increase retention with preference strength by adding $\Delta_i=2K|\lambda_i-1/2|$ ($K\geq0$):
\begin{equation}
r_i^{\ell_i}=
\begin{cases}
\Pi_a(\overline r_a+\Delta_i), & \ell_i=a,\\
\Pi_v(r_{i,0}^{v}+\Delta_i), & \ell_i=v,
\end{cases}
\label{eq:leading-modality-budget}
\end{equation}
where $\overline r_a$ is the base audio retention rate and $\Pi_a$ clips the audio token retention rate to its allowed range. These rates define each chunk's initial leading-modality retention demand.

\subsection{Budget-Constrained Semantic Compression}
Given the modality preference and initial leading-modality retention demand from TEGB, BCSC compresses the selected leading modality before its follower in each paired chunk.

\noindent\textbf{Video-led compression.}
BCSC recursively partitions each spatial token grid into regions represented by mean features, then merges similar, overlapping regions across adjacent time units to construct a candidate token mask.
The semantic weight $\gamma_v\geq0$ raises spatial and temporal similarity thresholds for query-relevant regions, encouraging finer partitions and more conservative merging.
Query-guided selection adjusts this mask to TEGB's target retention rate $r_i^v$; the retained visual fraction $\widehat r_i^v$ calibrates the audio target:
\begin{equation}
\check r_i^a=\Pi_a\!\left(
\overline r_a+
\beta(\widehat r_i^v-\overline r_v)
\right),
\label{eq:video-to-audio-retention}
\end{equation}
where $\beta\in[0,1]$ controls cross-modal coupling in both directions. Let $\mathbf a_{i,j}$ denote the embedding of the $j$-th audio token in chunk $i$. BCSC combines encoder-attention importance $h_{i,j}$ and query relevance $e_{i,j}^a=\cos(\mathbf q,\mathbf a_{i,j})$ into a selection score:
\begin{equation}
o_{i,j}^a=\widehat h_{i,j}+\gamma_a\widehat e_{i,j}^a,
\label{eq:audio-selection}
\end{equation}
where $\widehat h_{i,j}$ and $\widehat e_{i,j}^a$ are the within-chunk min-max normalized attention and query relevance scores, respectively; $\gamma_a\geq0$ controls semantic guidance. BCSC selects audio tokens in descending $o_{i,j}^a$ order at rate $\check r_i^a$ and adds regularly spaced context anchors from the remainder. Remaining tokens are assigned to their nearest anchors by cosine similarity. Within groups, tokens are ranked by maximum cosine similarity to retained visual tokens. Each anchor is averaged with a softmax-weighted aggregate of its top-ranked tokens.

\noindent\textbf{Audio-led compression.}
BCSC first retains audio tokens with the largest scores from Eq.~\eqref{eq:audio-selection}, using TEGB's initial audio target retention rate $r_i^a$ to determine their number. It adds regularly spaced context anchors from the remainder, assigns residual tokens to their most similar anchors, and merges top-ranked tokens within each group into these anchors using softmax-normalized visual similarities. Both residual token ranking and merging weights use all original visual tokens in the paired chunk, since video compression follows the audio step. The actual retained audio fraction $\widehat r_i^a$, including context anchors, calibrates the target visual retention rate:
\begin{equation}
\check r_i^v=\Pi_v\!\left(
\overline r_v+
\beta(\widehat r_i^a-\overline r_a)
\right).
\label{eq:audio-to-video-retention}
\end{equation}
BCSC then applies spatial and temporal masking, followed by query-guided visual token selection at this mapped rate.

\section{Experiments}

\begin{table*}[t]
  \centering
  \captionsetup{skip=2pt}
  \caption{Comparison with token compression methods across different omni-models.}
  \label{tab:main_results}

  \fontsize{8}{9}\selectfont
  \renewcommand{\arraystretch}{0.9}

  \setlength{\tabcolsep}{8.1pt}

  \begin{tabular}{@{} l l cc ccccc }
    \toprule

    Model
    & Method
    & \makecell{Retained Ratio}
    & \makecell{FLOPs Ratio}
    & WorldSense
    & AVUTBench
    & VideoMME
    & DailyOmni
    & Avg. \\
    \midrule


    \rowcolor{gray!40}
    \cellcolor{white}
    \multirow[c]{6}{*}{\makecell[l]{Qwen2.5-\\Omni-7B}}
    & Full Tokens
    & 100\% & 100\%
    & 46.8 & 64.5 & 66.0 & 62.9 & 100\% \\

    & Random
    & 55\% & 48\%
    & 43.6 & 61.0 & 65.4 & 58.6 & 95.0\% \\

    & DyCoke (V\&A)
    & 50\% & 44\%
    & 44.6 & 62.0 & 65.5 & 55.8 & 94.8\% \\

    & FlashVID
    & 45\% & 39\%
    & \underline{46.0} & \underline{63.0} & 65.9 & 57.0 & 96.6\% \\

    & OmniZip
    & 45\% & 39\%
    & 45.9 & \underline{63.0} & \textbf{66.1} & \underline{59.5}
    & \underline{97.6\%} \\

    \rowcolor{gray!15}
    \cellcolor{white}
    & OmniRoute (Ours)
    & 45\% & 39\%
    & \textbf{46.9} & \textbf{63.4} & \underline{66.0}
    & \textbf{60.6} & \textbf{98.7\%} \\

    \midrule


    \rowcolor{gray!40}
    \cellcolor{white}
    \multirow[c]{6}{*}{\makecell[l]{Qwen2.5-\\Omni-3B}}
    & Full Tokens
    & 100\% & 100\%
    & 46.4 & 62.2 & 62.6 & 62.0 & 100\% \\

    & Random
    & 55\% & 45\%
    & 42.8 & 58.7 & 61.1 & 54.8 & 93.2\% \\

    & DyCoke (V\&A)
    & 50\% & 40\%
    & 44.0 & 60.7 & 61.6 & 54.4 & 94.6\% \\

    & FlashVID
    & 45\% & 37\%
    & \underline{45.4} & 59.6 & 61.8 & 57.0 & 96.1\% \\

    & OmniZip
    & 45\% & 36\%
    & 45.2 & \underline{61.3} & \textbf{62.6}
    & \underline{58.6} & \underline{97.6\%} \\

    \rowcolor{gray!15}
    \cellcolor{white}
    & OmniRoute (Ours)
    & 45\% & 36\%
    & \textbf{45.8} & \textbf{61.7} & \textbf{62.6}
    & \textbf{58.7} & \textbf{98.1\%} \\

    \midrule


    \rowcolor{gray!40}
    \cellcolor{white}
    \multirow[c]{6}{*}{OmniVinci}
    & Full Tokens
    & 100\% & 100\%
    & 49.2 & 67.4 & 68.6 & 64.9 & 100\% \\

    & Random
    & 55\% & 49\%
    & 47.0 & 63.1 & 66.3 & 56.4 & 93.2\% \\

    & DyCoke (V\&A)
    & 50\% & 44\%
    & \underline{47.6} & 63.4 & 68.0 & 58.1 & 94.9\% \\

    & FlashVID
    & 50\% & 44\%
    & 44.7 & \underline{64.0} & 67.6 & 60.2 & 94.3\% \\

    & OmniZip
    & 45\% & 39\%
    & 47.1 & 63.7 & \textbf{68.4} & \underline{62.7}
    & \underline{96.6\%} \\

    \rowcolor{gray!15}
    \cellcolor{white}
    & OmniRoute (Ours)
    & 45\% & 38\%
    & \textbf{47.7} & \textbf{64.9} & \underline{68.2}
    & \textbf{64.1} & \textbf{97.9\%} \\

    \bottomrule
  \end{tabular}
  \vspace{-4mm}
\end{table*}

\begin{table*}[t]
  \vspace{1.5mm}

  \centering

  \begin{minipage}[t]{0.49\textwidth}
    \vspace{0pt}
    \centering

    \captionsetup{skip=2pt}

    \caption{
      Efficiency on WorldSense with Qwen2.5-Omni-7B.
      Mem.\ denotes peak GPU memory.
    }
    \label{tab:efficiency-comparison}

    \fontsize{8}{9}\selectfont
    \renewcommand{\arraystretch}{0.9}
    \setlength{\tabcolsep}{4.9pt}

    \begin{tabular}{lcccc}
      \toprule
      Method
      & \makecell{Mem. $\downarrow$}
      & \makecell{Prefill (ms) $\downarrow$}
      & Acc. $\uparrow$
      & \makecell{Latency (s) $\downarrow$} \\
      \midrule

      \rowcolor{gray!40}
      Full Tokens (100\%)
      & 44 G
      & 2371 (1.00$\times$)
      & \underline{46.8}
      & 10.99 (1.00$\times$) \\

      DyCoke (50\%)
      & 36 G
      & 1386 (1.71$\times$)
      & 44.6
      & 8.59 (1.28$\times$) \\

      FlashVID (45\%)
      & 35 G
      & 1073 (2.21$\times$)
      & 46.0
      & 9.63 (1.14$\times$) \\

      OmniZip (45\%)
      & \underline{32 G}
      & \textbf{894 (2.65$\times$)}
      & 45.9
      & \textbf{7.99 (1.38$\times$)} \\

      \rowcolor{gray!15}
      Ours (45\%)
      & \textbf{27 G}
      & \underline{936 (2.53$\times$)}
      & \textbf{46.9}
      & \underline{8.12 (1.35$\times$)} \\

      \bottomrule
    \end{tabular}

  \end{minipage}
  \hfill
  \begin{minipage}[t]{0.49\textwidth}
    \vspace{0pt}
    \centering

    \captionsetup{
      width=0.96\linewidth,
      skip=2pt
    }

    \caption{
      Ablation of OmniRoute's core components on WorldSense
      with Qwen2.5-Omni-7B.
    }
    \label{tab:module-ablation}

    \fontsize{8}{9}\selectfont
    \renewcommand{\arraystretch}{0.9}
    \setlength{\tabcolsep}{1mm}

    \begin{tabular*}{0.96\linewidth}{
        @{\extracolsep{\fill}}
        lcccc
        @{}
      }
      \toprule
      Setting
      & \makecell{$m_i$,$n_i$}
      & Routing
      & \makecell{Retained\\Ratio}
      & Acc. \\
      \midrule
      
      Full OmniRoute
      & \cmark
      & Dynamic
      & 45\%
      & \textbf{46.9} \\

      w/o $m_i$/$n_i$
      & \xmark
      & Dynamic
      & 45\%
      & 46.4 \\

      Audio-led only
      & \cmark
      & Audio-led
      & 45\%
      & 45.9 \\

      Video-led only
      & \cmark
      & Video-led
      & 45\%
      & 45.2 \\

      \bottomrule
    \end{tabular*}

  \end{minipage}

  \vspace{-2mm}
\end{table*}

\subsection{Experiment Setup}
\noindent\textbf{Datasets.} We evaluate OmniRoute on four representative benchmarks: WorldSense~\cite{hong2025worldsense}, covering audio--visual understanding across diverse real-world domains; VideoMME~\cite{fu2025video}, spanning short, medium, and long videos; AVUTBench~\cite{yang2025audio}, emphasizing audio-centric video understanding; and DailyOmni~\cite{zhou2025daily}, focusing on audio--visual reasoning in everyday scenarios.

\noindent\textbf{Implementation Details.} We evaluate our approach on Qwen2.5-Omni 3B/7B~\cite{xu2025qwen25omnitechnicalreport} and OmniVinci~\cite{ye2025omnivinci} using a single NVIDIA L20 GPU (48\,GB). For Qwen2.5-Omni, we uniformly sample videos at a target rate of 2 fps with a maximum of 768 frames per video. For OmniVinci, we uniformly sample 128 frames per video. Across all experiments, we set the spatial and temporal similarity thresholds to $\eta_s=0.82$ and $\eta_t=0.58$, respectively, the routing temperature to $T=0.03$, and the context-anchor ratio to $0.05$. The default semantic weights are $\gamma_v=0.21$ and $\gamma_a=0.09$, with a shared cross-modal coupling strength of $\beta=0.50$. Repeated OmniRoute evaluations under identical settings yielded identical accuracy.

\noindent\textbf{Baselines.} We use full-token inference as the uncompressed reference and random pruning (Random) as a control baseline to evaluate token selection effectiveness. We also compare with DyCoke (V\&A)~\cite{tao2025dycoke}, with its temporal compression module applied to both video and audio tokens; FlashVID~\cite{fan2026flashvid}, a strong visual token compression method for video LLMs; and OmniZip~\cite{tao2025omnizip}, a strong compression method specifically designed for Omni-LLMs.

\subsection{Main Results}
\noindent\textbf{Comparison with State-of-the-Art Methods.} As shown in Table~\ref{tab:main_results}, OmniRoute achieves the highest reported average normalized accuracy across Qwen2.5-Omni-7B/3B and OmniVinci on audio-visual understanding tasks. At $45\%$ token retention, it preserves $97.9$--$98.7\%$ of full-token accuracy on average across four benchmarks while reducing FLOPs by $61$--$64\%$, outperforming Random and DyCoke despite their higher token retention. This favorable accuracy--efficiency trade-off is further supported by gains of $1.1$, $0.5$, and $1.3$ percentage points over OmniZip at matched retention on Qwen2.5-Omni-7B/3B and OmniVinci, respectively. Notably, on WorldSense, OmniRoute achieves $46.9\%$ accuracy with Qwen2.5-Omni-7B at $45\%$ token retention, surpassing the full-token baseline.

\noindent\textbf{Efficiency Analyses.} On WorldSense with Qwen2.5-Omni-7B (Table~\ref{tab:efficiency-comparison}), OmniRoute achieves the lowest peak memory and highest accuracy among the compared methods at 45\% token retention. It reduces peak GPU memory by $38.6\%$ (44 to 27\,GB), with $2.53\times$ prefill and $1.35\times$ end-to-end speedups over full-token inference, outperforming DyCoke and FlashVID in both memory usage and latency. At matched retention, OmniRoute achieves prefill and end-to-end latencies comparable to those of OmniZip while reducing peak GPU memory by 5\,GB and improving accuracy by 1.0 percentage point. These results demonstrate improved inference efficiency while preserving predictive accuracy. 

\begin{figure}[t]
  \centering
  \captionsetup{skip=2pt}
  \includegraphics[width=\columnwidth]
  {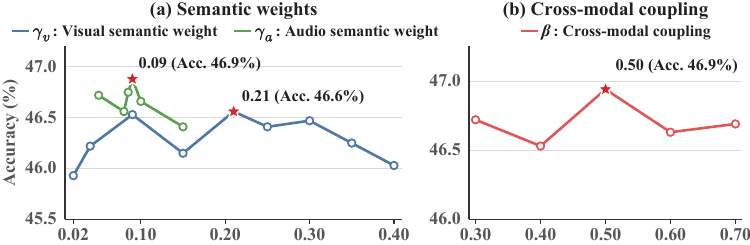}
  \vspace{-3mm}
  \caption{Hyperparameter analysis on WorldSense: modality-specific weights $\gamma_v$ and $\gamma_a$, and cross-modal coupling $\beta$.}
  \label{fig:hyperparameter-ablation}
  \vspace{-3mm}
\end{figure}

\subsection{Ablation Study}

\noindent\textbf{Ablation of Core Components.} Table~\ref{tab:module-ablation} examines budget adjustment and modality routing on WorldSense with Qwen2.5-Omni-7B at 45\% token retention. Disabling TEGB's within- and across-chunk variation signals ($m_i$ and $n_i$), while retaining dynamic routing, lowers accuracy from 46.9\% to 46.4\%. With these signals retained, fixed audio-led and video-led compression reduce accuracy by 1.0 and 1.7 percentage points, respectively. These results highlight the benefits of temporally adaptive budgeting and modality routing for preserving audio--visual understanding under token compression.

\noindent\textbf{Analysis of Compression Hyperparameters.} Fig.~\ref{fig:hyperparameter-ablation} examines sensitivity to semantic guidance and cross-modal coupling on WorldSense. We first vary $\gamma_v$, which adjusts query-dependent spatial and temporal similarity thresholds, with accuracy peaking at $\gamma_v=0.21$ (46.6\%). Fixing $\gamma_v=0.21$, we then vary $\gamma_a$, which controls the contribution of query relevance relative to encoder attention; accuracy peaks at $\gamma_a=0.09$ (46.9\%). Both sweeps exhibit non-monotonic trends, indicating that stronger semantic guidance does not necessarily improve accuracy. For cross-modal coupling, $\beta=0.50$ yields the highest tested accuracy (46.9\%), with lower scores at both smaller and larger coupling strengths. Within the tested ranges, these results favor moderate semantic guidance and cross-modal coupling for preserving audio--visual understanding under compression. Accordingly, we adopt these values as the default hyperparameter settings.

\noindent\textbf{Effect of Token Retention Ratio.}
Table~\ref{tab:retained-ratio-ablation} evaluates OmniRoute under different token budgets on WorldSense with Qwen2.5-Omni-7B. As token retention decreases from 45\% to 30\%, accuracy declines from 46.9\% to 45.6\%, while FLOPs and prefill time decrease by 54\% and 40\%, respectively. End-to-end latency decreases from 8.12 to 7.59\,s. These results show that OmniRoute offers a flexible accuracy-efficiency trade-off, achieving faster inference under tighter token budgets with limited accuracy degradation.

\begin{table}[t]
  \centering

  \captionsetup{skip=2pt}

  \caption{
    Performance of OmniRoute under different retained ratios
    on WorldSense with Qwen2.5-Omni-7B.
  }
  \label{tab:retained-ratio-ablation}

  \fontsize{8}{9}\selectfont
  \renewcommand{\arraystretch}{0.9}
  \setlength{\tabcolsep}{1mm}

  \begin{tabular*}{\linewidth}{
      @{\extracolsep{\fill}}
      lccccc
      @{}
    }
    \toprule
    \makecell{Retained\\Ratio}
    & \makecell{FLOPs\\Ratio}
    & \makecell{Mem.  $\downarrow$}
    & \makecell{Prefill (ms) $\downarrow$}
    & \makecell{Acc. $\uparrow$}
    & \makecell{Latency (s) $\downarrow$} \\
    \midrule

    45\%
    & 39\%
    & 27 G
    & 936
    & 46.9
    & 8.12 \\

    40\%
    & 31\%
    & 27 G
    & 788
    & 46.3
    & 8.00 \\

    35\%
    & 26\%
    & 27 G
    & 640
    & 46.1
    & 7.73 \\

    30\%
    & 18\%
    & 27 G
    & 558
    & 45.6
    & 7.59 \\

    \bottomrule
  \end{tabular*}
  \vspace{-3mm}

\end{table}

\section{Conclusion}
We introduce OmniRoute, a training-free, two-stage token compression framework for Omni-LLMs. It combines temporal evidence-guided budgeting with modality-led semantic compression, calibrating follower budgets using actual leading-modality retention. Experiments on four benchmarks across three models demonstrate that OmniRoute achieves a better accuracy-efficiency trade-off than competitive baselines. Despite these gains, OmniRoute still relies on manually configured modality-specific semantic weights and cross-modal coupling strength. Future work will explore adapting these parameters to target token budgets to reduce manual tuning. We will also optimize the implementation to lower inference latency.

\bibliographystyle{IEEEbib}
\bibliography{strings,refs}

\end{document}